\documentclass[letterpaper]{article} 
\usepackage[preprint]{aaai2027} 
\usepackage[hyphens]{url} 
\usepackage{graphicx} 
\usepackage{natbib} 
\usepackage{caption} 
\usepackage{booktabs}
\usepackage{amsmath}
\usepackage{newfloat}
\usepackage{float}
\usepackage{listings}
\usepackage{xcolor}

\DeclareCaptionStyle{ruled}{labelfont=normalfont,labelsep=colon,
  strut=off}
\floatstyle{ruled}
\newfloat{listing}{tb}{lst}{}
\floatname{listing}{Listing}

\newcommand{\twohead}[2]{\shortstack{\strut #1\\\strut #2}}

\title{HierDoc: Hierarchical Page-to-Region Evidence Routing for Long-Document Visual Question Answering}
\author{%
    \makebox[\textwidth][c]{Rongjian Gu\textsuperscript{\rm 1}\qquad Wengang Zhou\textsuperscript{\rm 1}\qquad Junyu Xiong\textsuperscript{\rm 1}\qquad Yonghui Wang\textsuperscript{\rm 1}}\\
    \makebox[\textwidth][c]{Bing Yin\textsuperscript{\rm 2}\qquad Bei Wang\textsuperscript{\rm 3}\qquad Houqiang Li\textsuperscript{\rm 1}}%
}
\affiliations{%
    \makebox[\textwidth][c]{\textsuperscript{\rm 1}University of Science and Technology of China\qquad \textsuperscript{\rm 2}iFLYTEK Research}\\
    \makebox[\textwidth][c]{\textsuperscript{\rm 3}Anhui Transport Consulting \& Design Institute Co. Ltd., Hefei, China}\\
    \makebox[\textwidth][c]{gurongjian@mail.ustc.edu.cn, xiongjyu@mail.ustc.edu.cn, zhwg@ustc.edu.cn}%
}

\begin{document}

\maketitle

\begin{abstract}
Multi-page document visual question answering requires locating sparse evidence at both the page and region levels.
Existing approaches typically emphasize one level over the other: page-centric methods focus on page acquisition, with region operations serving mainly as navigation aids, whereas region-centric methods assume that the relevant pages have already been supplied.
Consequently, page and region selection remain disconnected rather than forming successive evidence decisions.
We propose HierDoc, a hierarchical evidence-routing framework that formulates long-document evidence acquisition as two-stage set prediction from pages to regions.
A page policy selects evidence pages from the full document; these pages are then parsed for semantic elements, after which a region policy selects the elements passed to a downstream answer model.
Both answer-agnostic policies are optimized with stage-wise GRPO using granularity-specific structured-set rewards.
The answer model receives selected full pages together with selected region crops and OCR or table text, preserving global context while emphasizing fine-grained evidence.
Across the evaluated benchmarks, HierDoc achieves state-of-the-art or competitive performance among open-weight systems, improving LongDocURL by 16.87\% relative to the strongest reported open-weight baseline.
Controlled ablations further show that selected regional evidence improves the page-only system in accuracy and F1 by 5.51\% and 4.82\%, respectively.
These results demonstrate the benefit of organizing coarse page routing and fine-grained region routing as successive, separately optimized stages of a unified evidence-acquisition process.
\end{abstract}

\section{Introduction}

Document visual question answering is moving from single-page reading toward multi-page and long-document reasoning, where answer evidence is often sparse across pages and visual elements.
DocVQA established question answering over individual document images \citep{Mathew_2021_WACV}, while MP-DocVQA, SlideVQA, and DUDE extended the task to collections of related pages \citep{tito2023hierarchical,tanaka2023slidevqa,Van_Landeghem_2023_ICCV}.
More recent benchmarks such as MMLongBench-Doc and LongDocURL further increase document length, visual diversity, and evidence-localization demands \citep{ma2024mmlongbenchdoc,deng-etal-2025-longdocurl}.
As the context expands, the central challenge shifts from understanding a supplied page to acquiring the right evidence before reasoning: a system must identify relevant pages and then locate relevant semantic regions within them.

Existing approaches typically operate primarily at either the page level or the within-page region level.
Visual retrievers such as ColPali and VisRAG use rendered pages as retrieval units \citep{faysse2025colpali,yu2025visrag}, and DocR1 makes evidence-page selection trainable through page-guided GRPO \citep{xiong2026docr1}.
Agentic systems such as Doc-$V^*$ extend page acquisition beyond one-shot retrieval \citep{zheng-etal-2026-doc}.
Across these page-oriented approaches, fine-grained evidence is therefore not formulated as a separately optimized region-set action.
At the within-page level, AgenticOCR learns query-driven zoom-and-OCR operations over pages supplied by an external page retriever, making within-page extraction explicit while taking page acquisition as an upstream input \citep{wang2026agenticocr}.
DocLens most directly connects these granularities: its Page Navigator retrieves relevant pages, and its Element Localizer applies query-agnostic layout detection and exhaustively crops detected elements for downstream answering \citep{zhu-etal-2026-doclens}.
Although this establishes a full-document-to-element path, it neither learns query-conditioned region-subset selection nor optimizes page and region acquisition as separate evidence policies.
Consequently, existing methods lack a unified formulation of page and region routing as successive, answer-agnostic evidence-set policies optimized at their respective granularities.

HierDoc addresses this gap with a hierarchical evidence-routing framework that casts evidence acquisition as successive page and region-level set prediction.
Given a question and a rendered document, a page policy first returns a structured set of evidence pages.
Selected pages are then parsed into candidate semantic regions with bounding boxes, element types, and recognized textual content \citep{niu-etal-2026-mineru2}.
A region policy conditions on the question, selected-page images, and parser candidates to return a structured set of evidence-region IDs.
A separate answer model receives the selected pages together with the selected region crops and textual metadata, retaining global layout context while emphasizing local evidence.
Each routing policy is answer-agnostic, terminates at an evidence set, and is optimized independently with stage-wise GRPO \citep{shao2024deepseekmath}.
Granularity-specific structured-set rewards let the page policy balance evidence coverage and context volume while the region policy selects precise semantic elements and penalizes surplus actions.
Bounding-box supervision is mapped to parser-native region IDs to form a discrete semantic action space; at inference time, the two policies compose into one page-to-region-to-answer path.
Figure~\ref{fig:hierdoc-inference} summarizes this inference path and the evidence passed between stages.

Experiments across multiple multi-page and long-document VQA benchmarks test end-to-end answering, page routing, and evidence composition.
Across the reported evaluations, HierDoc achieves state-of-the-art or competitive performance among open-weight systems, including a 16.87\% relative accuracy improvement over the strongest reported open-weight baseline on LongDocURL.
Controlled evidence-composition ablations further show that selected regional evidence improves the page-only system by 5.51\% in accuracy and 4.82\% in F1, supporting the complementary roles of global page context and fine-grained region routing.

Our contributions are threefold:
\begin{itemize}
    \item We introduce HierDoc, a hierarchical evidence-routing framework that represents page and semantic-region acquisition as successive, answer-agnostic structured-set policies, making both routing decisions explicit and independently analyzable.
    \item We develop a stage-wise training strategy with granularity-specific set rewards and a parser-native semantic action space that maps spatial supervision to discrete region decisions.
    \item We evaluate HierDoc across multi-page and long-document VQA benchmarks and conduct controlled routing and evidence-composition studies, demonstrating strong open-weight performance and the complementary value of page context and selected regional evidence.
\end{itemize}

\section{Related Work}

\subsection{MLLMs for Document VQA}

Document VQA requires joint reasoning over text, layout, and visual structure.
Beyond DocVQA, InfographicVQA, ChartQA, and TAT-DQA emphasize infographic, chart, and financial table--text reasoning, respectively \citep{Mathew_2021_WACV,Mathew_2022_WACV,masry-etal-2022-chartqa,zhu2022tatdqa}.
Multi-page benchmarks such as MP-DocVQA, SlideVQA, and DUDE further require cross-page evidence aggregation \citep{tito2023hierarchical,tanaka2023slidevqa,Van_Landeghem_2023_ICCV}, while MMLongBench-Doc and LongDocURL extend evaluation to long, visualization-rich PDFs with explicit localization demands \citep{ma2024mmlongbenchdoc,deng-etal-2025-longdocurl}.
General-purpose MLLMs, including Qwen2.5-VL, GLM-4.5V, and InternVL3.5, directly process document images with high-resolution perception and multimodal reasoning \citep{bai2025qwen25vl,glm-v2025glm45v,wang2025internvl35}.
Document-specialized mPLUG-DocOwl2 compresses pages into compact visual tokens for OCR-free multi-page understanding, while DeepSeek-OCR explores optical compression of long textual contexts \citep{hu-etal-2025-mplug,wei2025deepseekocr}.
Long documents therefore motivate explicit evidence routing across pages and regions, which motivates HierDoc's hierarchical evidence-routing design.

\subsection{Evidence Routing across Pages and Regions}

Visual document retrieval provides a natural first step toward evidence acquisition.
For visually rich documents, ColPali represents page images with multi-vector embeddings \citep{faysse2025colpali}.
M3DocRAG couples multimodal page retrieval with an MLLM answerer, SV-RAG adapts an MLLM with separate retrieval and answering modules, and MoLoRAG augments page selection with graph-based logical links \citep{cho2024m3docrag,chen2025svrag,wu-etal-2025-molorag}.
These approaches demonstrate the value of query-conditioned page selection, but the retrieved object remains an entire page and fine-grained evidence is left implicit.
DocR1 uses Evidence Page-Guided GRPO to reward page judgments together with reasoning format and answer accuracy \citep{xiong2026docr1}.
This establishes page evidence as an optimizable behavior rather than fixed preprocessing and provides precedent for learning evidence selection with verifiable set rewards.

Finer-grained methods localize evidence within document pages.
RegionRAG learns localized visual representations and groups salient patches into semantic regions before generation \citep{li2026regionrag}.
AgenticOCR learns query-conditioned zoom-and-OCR operations over externally retrieved pages \citep{wang2026agenticocr}.
These designs provide within-page localization but generally remain separate from upstream page retrieval.
HierDoc connects page acquisition and semantic localization by selecting a discrete evidence set from parser-native regions conditioned on the pages chosen upstream.

\subsection{Reinforcement-Learned and Agentic Document Reasoning}

Agentic document systems extend static retrieval by allowing models to seek and combine evidence during reasoning.
In visual document understanding, ViDoRAG coordinates retrieval, exploration, summarization, and reflection; MDocAgent combines specialized text and image agents \citep{wang-etal-2025-vidorag,han2025mdocagent}.
DocLens composes page navigation, element localization, answer sampling, and adjudication in a tool-augmented multi-agent workflow \citep{zhu-etal-2026-doclens}.

Recent work also optimizes evidence-seeking trajectories with reinforcement learning.
Doc-$V^*$ alternates retrieval, page fetching, and answering via structured memory, VRAG-RL learns iterative coarse-to-fine visual search, and HIEVI-RAG combines hierarchical question decomposition with GRPO-trained page verification \citep{zheng-etal-2026-doc,wang2025vragrl,xiong2026hierarchical}.
These closed-loop designs couple evidence discovery with answer-oriented reasoning.
HierDoc adopts a complementary factorization: page and semantic-region acquisition are standalone structured-set policies, and answer generation begins only after both routing stages terminate.

\begin{figure*}[t]
\centering
\includegraphics[width=\textwidth]{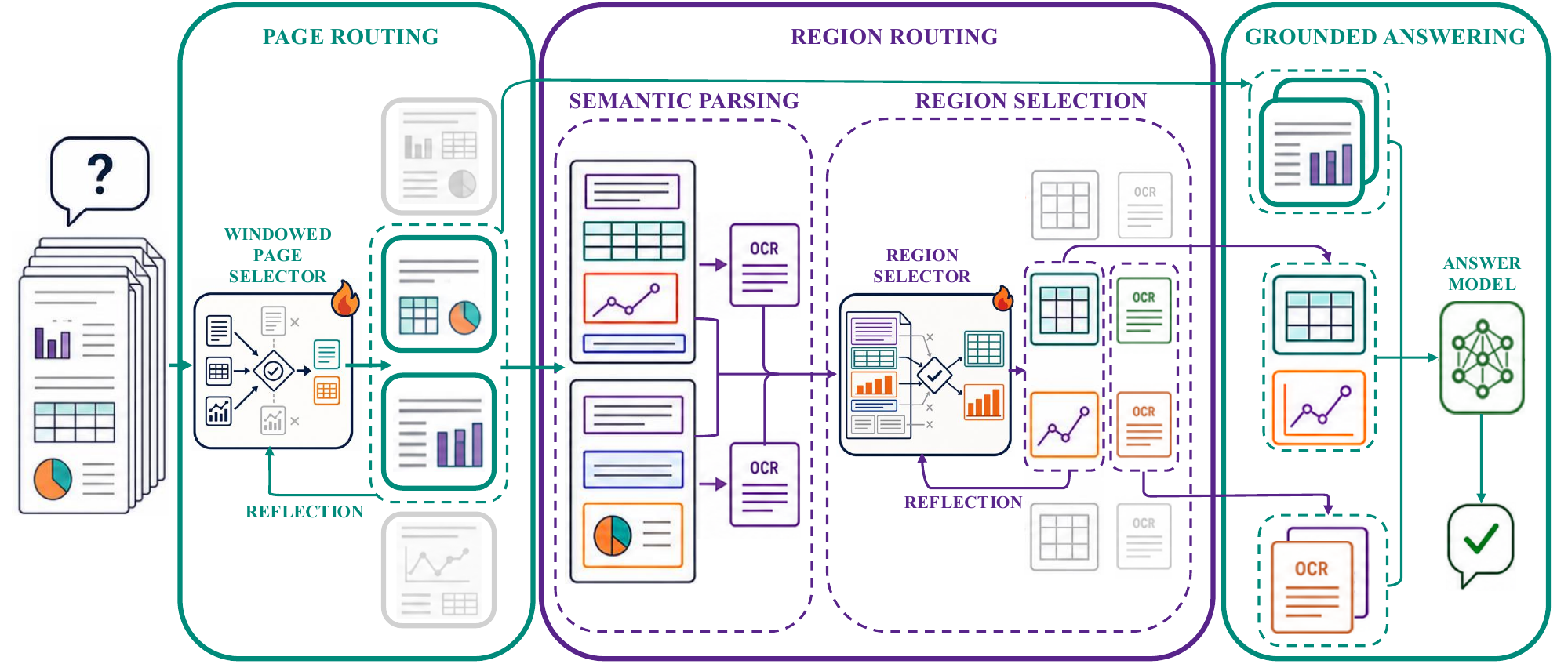}
\caption{Overview of HierDoc inference. A trainable page policy selects evidence pages from fixed-size windows and routes them both to semantic parsing and directly to grounded answering. MinerU2.5-Pro converts the selected pages into typed semantic regions with bounding boxes and OCR or table metadata, after which a trainable region policy retains the relevant region crops and associated text. The answer model jointly consumes the selected full pages, region crops, and textual metadata. Flame badges denote trainable selectors.}
\label{fig:hierdoc-inference}
\end{figure*}

\section{Method}

\subsection{Overview and Problem Formulation}

Given a question $q$ and a document $D=(I_1,\ldots,I_N)$ of rendered page images, HierDoc produces an answer in three stages.
First, \emph{page routing} selects the pages likely to contain answer evidence.
Second, \emph{region routing} parses those pages into semantic regions and retains the regions most relevant to the question.
Finally, \emph{grounded answering} combines the selected full pages with the selected region crops and their textual metadata to generate the answer.
Figure~\ref{fig:hierdoc-inference} presents the overall workflow of the HierDoc framework.

Let $D_{\hat P}=\{I_p:p\in\hat P\}$ denote the selected page images, where $\hat P\subseteq\{1,\ldots,N\}$ contains 1-based physical page indices.
MinerU2.5 \citep{niu-etal-2026-mineru2} parses these images into the candidate set
\[
\mathcal C(\hat P)=
\{r_j=(a_j,p_j,b_j,u_j,t_j)\}_{j=1}^{M},
\]
where $a_j$ is a sample-local action alias, $p_j\in\hat P$ is the physical page index, $b_j=(x_1,y_1,x_2,y_2)$ is a pixel-space bounding box, $u_j$ is a semantic type, and $t_j$ is OCR or table text.
For a selected region set $\hat R\subseteq\mathcal C(\hat P)$, the selected crops and their associated metadata are denoted as
\[
\mathcal E_{\hat R}=
\{(\operatorname{crop}(I_{p_j},b_j),r_j):r_j\in\hat R\}.
\]
To summarize, the inputs and outputs of the three stages are
\[
\begin{aligned}
\pi_p &: (q,D) \mapsto \hat P,\\
\pi_r &: (q,D_{\hat P},\mathcal C(\hat P)) \mapsto \hat R,\\
F &: (q,D_{\hat P},\mathcal E_{\hat R}) \mapsto \hat y.
\end{aligned}
\]

\subsection{Page Routing}

\paragraph{Windowed Page Selection}
For a candidate page set, the page policy returns exactly one XML action of the form \texttt{<evidence\_page>2,4</evidence\_page>} and is instructed not to answer the question.
An empty tag denotes an empty page set, and indices outside the presented candidates are invalid actions.
At inference time, a long document is partitioned into consecutive, non-overlapping windows $W_1,\ldots,W_K$, each containing at most $M_p$ pages, where $M_p$ denotes the page-window capacity.
The policy acts independently on each window, after which valid local predictions are deduplicated and combined by union:
\[
\hat P^{(0)}=\bigcup_{k=1}^{K}\pi_p(q,W_k).
\]
This construction lets the same set policy operate on documents longer than a single visual context while preserving physical page identities across windows.

After the window-level predictions are merged, heuristic reflection is triggered when $|\hat P^{(0)}|>\tau_p$, where $\tau_p$ denotes the page-reflection threshold.
The same page policy then reconsiders only the first-pass pages and returns a refined set $\hat P$ satisfying $|\hat P|<M_p$; otherwise $\hat P=\hat P^{(0)}$.
If the second pass does not produce an admissible action, the first-pass set is retained.
By pruning redundant pages accumulated across independent windows, reflection reduces the visual context passed to semantic parsing and grounded answering while requiring neither an additional policy nor a separate training objective.

\paragraph{Page-Policy Optimization}
To improve evidence coverage without forwarding excessive or malformed page sets, we optimize the page policy $\pi_p$ with Group Relative Policy Optimization (GRPO) \citep{shao2024deepseekmath}.
For each question--window prompt, $\pi_p$ samples $N$ candidate page-set actions, standardizes their rule-based rewards within the group, and uses the resulting relative advantages to update the policy under a clipped surrogate objective with KL regularization to the reference policy.
The key component is a structured reward that evaluates both the selected page set and its serialization.
Let $o_p$ be the serialized action for a candidate index set $W$, $\widetilde P$ its parsed indices, and $G_p$ the document-level gold page set.
We define $U_p=\widetilde P\setminus W$, the valid set $S_p=\widetilde P\cap W$, and the window-local target $G_{p,W}=G_p\cap W$.
The reward is
\[
\begin{aligned}
\mathcal R_p=\operatorname{clip}_{[0,1]}\!\Bigl(&
\lambda_{\mathrm{rec}}^p\,\operatorname{Rec}(S_p,G_{p,W})+
\lambda_{\mathrm{f1}}^p\,\operatorname{F1}(S_p,G_{p,W})\\
&+\lambda_{\mathrm{fmt}}^p\,\operatorname{Fmt}(o_p)
-\Omega_{\mathrm{inv}}(U_p)
\Bigr).
\end{aligned}
\]
We define $\Omega_{\mathrm{inv}}(U_p)=\min(0.20,0.05|U_p|)$.
The recall and F1 terms favor evidence coverage, while F1 also discourages forwarding the entire window.
The format term and invalid-index penalty enforce valid serialization, and clipping keeps the combined reward in $[0,1]$.

\subsection{Region Routing}

\paragraph{Semantic Parsing}
For each selected page, MinerU2.5-Pro decomposes the page into semantically typed units such as paragraphs, tables, figures, and captions.
Each unit retains its page index, pixel-space bounding box, semantic type, and available OCR or table text.
The resulting parser-native regions form a discrete action space, avoiding the need for the selector to generate free-form coordinates.
We assign consecutive aliases $a_j\in\{1,\ldots,M\}$ that are unique within each sample and shared across all of its selected pages.

\paragraph{Region Selection}
The region policy observes the question, selected pages with the aliases overlaid on their corresponding regions, and a candidate list formatted as \texttt{alias|type|OCR hint}.
The page images remain the authoritative visual input, while the recognized text serves as an auxiliary cue.
Following the same protocol as the page policy, $\pi_r$ outputs a single cross-page action such as \texttt{<evidence\_region>3,17</evidence\_region>} over the sample-local aliases; an empty tag denotes no selected region, and out-of-list aliases are invalid.

Let $\hat R^{(0)}$ denote the first-pass region set.
As in page routing, reflection is triggered only when $|\hat R^{(0)}|>\tau_r$, where $\tau_r$ denotes the region-reflection threshold.
The same policy then reconsiders the first-pass candidates; otherwise $\hat R=\hat R^{(0)}$, and a failed refinement falls back to the first-pass set.
Unlike page reflection, region reflection imposes no fixed output limit and instead seeks the smallest sufficient subset $\hat R$, reducing the crop and OCR context passed to grounded answering.

\paragraph{Region-Policy Optimization}
To improve fine-grained evidence coverage without forwarding all parser candidates, we optimize $\pi_r$ with a similar GRPO procedure to the page policy, with independent training prompts and a region-specific reward.
Training supervision is originally provided as evidence boxes, whereas the policy acts on parser-native aliases.
To map the original annotations to these aliases, we compare each gold evidence box $g$ with every parser candidate box $b_j$ in the same page coordinate system.
Candidate $j$ is assigned to the gold region set when
\[
\operatorname{IoU}(g,b_j)\geq0.10
\quad\lor\quad
\frac{|g\cap b_j|}{|b_j|}\geq0.80.
\]
The containment criterion captures semantic units lying inside a larger annotated evidence area even when their IoU is small.
All matched candidates are retained and merged across gold boxes to form the gold region-action set, which is then subjected to fine-grained quality filtering.
To provide document-consistent distractors, training prompts also include randomly sampled non-evidence pages from the same document; their parsed regions enter the candidate set but never the gold region set.

Let $o_r$ be a serialized region action, $\widetilde A_r$ its parsed aliases, $\mathcal A(\hat P)=\{a_j:r_j\in\mathcal C(\hat P)\}$ the candidate alias set, and $G_r\subseteq\mathcal A(\hat P)$ the mapped gold aliases.
We define $U_r=\widetilde A_r\setminus\mathcal A(\hat P)$ and $S_r=\widetilde A_r\cap\mathcal A(\hat P)$.
Analogous to the page reward, the region reward combines recall, F1, precision, format compliance, and an invalid-action penalty:
\[
\begin{aligned}
\mathcal R_r=\operatorname{clip}_{[0,1]}\!\Bigl(&
\lambda_{\mathrm{rec}}^r\,\operatorname{Rec}(S_r,G_r)
+\lambda_{\mathrm{f1}}^r\,\operatorname{F1}(S_r,G_r)\\
&+\lambda_{\mathrm{prec}}^r\,\operatorname{Prec}(S_r,G_r)
+\lambda_{\mathrm{fmt}}^r\,\operatorname{Fmt}(o_r)\\
&-\Omega_{\mathrm{inv}}(U_r)
\Bigr).
\end{aligned}
\]
We define $\Omega_{\mathrm{inv}}(U_r)=\min(0.50,0.10|U_r|)$.
The clipped reward is set to zero if the evidence-region tag is missing, capped at $0.25$ for a non-strict action, and capped at $0.10$ for an action containing any invalid alias.
The dominant F1 term balances evidence coverage and selection precision, while the explicit recall and precision terms further penalize missed evidence and unnecessary regions, respectively.

\subsection{Grounded Answering}

If the final page set is empty, the system returns \texttt{Not answerable} without invoking the remaining stages.
Otherwise, MinerU and the region policy operate on the selected pages.
For a non-empty region set, the answer model receives the clean selected page images, available selected-region crops, region metadata, and cleaned OCR or table text.
Full pages preserve layout, legends, and cross-element relations; crops increase the visual prominence of local evidence; and OCR supplies a compact textual view of the same units.
If the region set is empty, the crop and OCR channels are omitted and the answer model operates on the selected full pages alone.
The answer stage therefore uses region routing as local evidence augmentation while retaining the global context established by page routing.

\section{Experiments}

\begin{table*}[t]
\centering
\resizebox{\textwidth}{!}{%
\begin{tabular}{lcccccc}
\toprule
Method & Model Size & \twohead{MMLongBench-Doc}{(Acc.)} & \twohead{LongDocURL}{(Acc.)} & \twohead{SlideVQA}{(F1)} & \twohead{PaperTab}{(Acc.)} & \twohead{FetaTab}{(Acc.)} \\
\midrule
mPLUG-DocOwl2 & 8B & 13.40 & 5.30 & 27.80 & -- & -- \\
InternVL3 & 8B & 24.10 & 38.70 & 64.40 & -- & -- \\
Qwen2.5-VL & 7B & 28.00 & 32.90 & 55.20 & 12.70 & 32.90 \\
\midrule
M3DocRAG & 7B & 21.00 & 35.10 & 55.70 & 28.50 & 63.80 \\
MoLoRAG+ & 7B & 41.00 & 51.90 & -- & \underline{31.00} & \underline{69.20} \\
CogDoc & 7B & 33.00 & -- & 67.90 & -- & -- \\
Doc-$V^*$ & 7B & 42.10 & 56.30 & \textbf{77.20} & -- & -- \\
MDocAgent & 7B & 38.50 & 46.90 & -- & 30.00 & 66.30 \\
\midrule
\textbf{HierDoc (Qwen2.5-VL)} & \textbf{7B} & \underline{51.29} & \underline{61.65} & 71.88 & 30.90 & 58.55 \\
\textbf{HierDoc (Qwen3-VL)} & \textbf{8B} & \textbf{53.62} & \textbf{65.80} & \underline{76.45} & \textbf{42.49} & \textbf{69.50} \\
\bottomrule
\end{tabular}%
}
\caption{Main results on five long-document VQA benchmarks.
Each dataset uses the metric shown in its column.
All scores are reported to two decimal places.
The best and second-best directly comparable results in each column are shown in bold and underlined, respectively.}
\label{tab:main-results}
\end{table*}

\subsection{Experiment Setup}

\paragraph{Datasets and Metrics}

We evaluate HierDoc on five multi-page and long-document benchmarks.
MMLongBench-Doc contains visually rich long PDFs that require understanding text, tables, charts, and cross-page evidence; we follow its official 1,082-question protocol and report accuracy \citep{ma2024mmlongbenchdoc}.
LongDocURL targets understanding, numerical reasoning, and cross-element localization in long multimodal documents; we report the official average accuracy produced by its evaluator \citep{deng-etal-2025-longdocurl}.
SlideVQA evaluates question answering over complete slide decks, including single-hop, multi-hop, and numerical questions; we report answer F1 on the 2,215-question test split \citep{tanaka2023slidevqa}.
PaperTab and FetaTab are the table-oriented academic-paper and Wikipedia subsets of UDA-QA, with 393 and 1,023 test questions, respectively \citep{hui2024uda}.
Following common practice, we adopt an LLM-as-a-judge protocol for both datasets \citep{zheng2023judging}.
Specifically, GPT-5.5 assigns a binary correctness score to each prediction over the complete test splits.

\paragraph{Compared Methods}

We compare against the direct multimodal backbones InternVL3, mPLUG-DocOwl2, and Qwen2.5-VL \citep{zhu2025internvl3,hu-etal-2025-mplug,bai2025qwen25vl}.
The document-specific and retrieval-based baselines comprise M3DocRAG, MoLoRAG+, CogDoc, Doc-$V^*$, and MDocAgent \citep{cho2024m3docrag,wu-etal-2025-molorag,xu2025cogdoc,zheng-etal-2026-doc,han2025mdocagent}.

\paragraph{Implementation Details}

All experiments are conducted on 8 NVIDIA A100 GPUs.
The page and region selectors are initialized from Qwen3-VL-8B-Thinking and trained independently, while the final answer model is instantiated with either Qwen2.5-VL-7B-Instruct or Qwen3-VL-8B-Instruct \citep{bai2025qwen3vl,bai2025qwen25vl}.
The page-policy corpus contains 2,200 examples from MMDocIR's subset, which is annotated from DUDE and MP-DocVQA \citep{dong-etal-2025-mmdocir}.
The region corpus begins with 2,418 ViDoRe-v3 records; parser alignment and data auditing retain 1,022 clean records \citep{loison2026vidore}.
We optimize both selectors with GRPO for one epoch.
For the page-policy reward, we set $(\lambda_{\mathrm{rec}}^p,\lambda_{\mathrm{f1}}^p,\lambda_{\mathrm{fmt}}^p)=(0.40,0.45,0.15)$.
For the region-policy reward, we set $(\lambda_{\mathrm{rec}}^r,\lambda_{\mathrm{f1}}^r,\lambda_{\mathrm{prec}}^r,\lambda_{\mathrm{fmt}}^r)=(0.20,0.50,0.20,0.10)$.
The page and region policies use global prompt batch sizes of 32 and 16, respectively.
For each prompt, we sample $N=4$ rollouts.
Both policies use a learning rate of $1\times10^{-6}$ and low-variance KL regularization with a coefficient of 0.01.
Selector images have a maximum long edge of 1,024 pixels.
At inference time, we set the page-window capacity to $M_p=16$ and the reflection thresholds to $\tau_p=\tau_r=8$.

\subsection{Main Results}

\paragraph{End-to-End QA Performance}
HierDoc achieves the best reported performance on four of the five benchmarks and remains competitive on SlideVQA, as shown in Table~\ref{tab:main-results}.
With Qwen3-VL-8B-Instruct as the answer model, HierDoc delivers relative improvements of 27.36\% and 16.87\% over Doc-$V^*$ on MMLongBench-Doc and LongDocURL, respectively.
Relative to the strongest reported baselines, it further improves PaperTab and FetaTab by 37.06\% and 0.43\%, respectively.
On SlideVQA, the relative gap to Doc-$V^*$ is only 0.97\%.
Together, these results demonstrate broad effectiveness across long PDFs, table-oriented documents, and slide decks.

The performance gains are not solely attributable to the stronger Qwen3-VL answer model.
When paired with Qwen2.5-VL-7B-Instruct, HierDoc substantially outperforms the direct backbone across the evaluated benchmarks and, on MMLongBench-Doc and LongDocURL, also surpasses every competing method.
This result shows that the hierarchical evidence-routing pipeline transfers across answer backbones and remains effective without relying on the stronger answer model.

\begin{table*}[t]
\centering
\begin{tabular}{llcccc}
\toprule
Dataset & Page-retrieval method & Page P & Page R & Page F1 & Avg. pages \\
\midrule
MMLongBench-Doc & ColQwen + Qwen2.5-VL & -- & -- & 30.9 & 6.0 \\
MMLongBench-Doc & Doc-$V^*$ + ColQwen & -- & -- & 49.7 & 5.6 \\
MMLongBench-Doc & \textbf{HierDoc page policy} & \textbf{67.9} & \textbf{76.9} & \textbf{68.9} & \textbf{2.2} \\
\midrule
LongDocURL & Doc-$V^*$ \texttt{retrieve\_page} action & 32.7 & \textbf{83.4} & 44.4 & -- \\
LongDocURL & Doc-$V^*$ \texttt{fetch\_page} action & 36.6 & 37.3 & 31.9 & -- \\
LongDocURL & \textbf{HierDoc page policy} & \textbf{71.9} & 75.2 & \textbf{69.6} & \textbf{2.1} \\
\midrule
SlideVQA & Doc-$V^*$ \texttt{retrieve\_page} action & 39.0 & \textbf{95.7} & 54.1 & -- \\
SlideVQA & Doc-$V^*$ \texttt{fetch\_page} action & 81.2 & 70.9 & 72.9 & -- \\
SlideVQA & \textbf{HierDoc page policy} & \textbf{81.9} & 88.4 & \textbf{82.3} & \textbf{1.5} \\
\bottomrule
\end{tabular}
\caption{Page-level evidence retrieval.
Doc-$V^*$ reports its \texttt{retrieve\_page} and \texttt{fetch\_page} actions separately.}
\label{tab:page-retrieval}
\end{table*}

\begin{table}[t]
\centering
\small
\setlength{\tabcolsep}{2.5pt}
\begin{tabular}{@{}p{0.40\columnwidth}cccc@{}}
\toprule
Incremental configuration & \twohead{Selector}{F1} & \twohead{Avg.}{selected} & \twohead{QA}{Acc.} & \twohead{QA}{F1} \\
\midrule
\multicolumn{5}{@{}l}{\textit{Page routing (average selected pages)}} \\
Untrained backbone & 57.3 & 2.6 & 47.5 & 43.2 \\
\quad + GRPO training & 67.8 & 2.6 & 50.0 & 45.1 \\
\quad \textbf{+ Bounded reflection} & \textbf{68.9} & \textbf{2.2} & \textbf{50.8} & \textbf{45.6} \\
\midrule
\multicolumn{5}{@{}l}{\textit{Region routing (average selected regions)}} \\
Page-only (no region policy) & N/A & N/A & 50.8 & 45.6 \\
\quad + Untrained region policy & 46.9 & 1.1 & 51.9 & 45.8 \\
\quad + GRPO training & 46.9 & 3.7 & 52.7 & 46.9 \\
\quad \textbf{+ Bounded reflection} & \textbf{59.5} & \textbf{2.9} & \textbf{53.6} & \textbf{47.8} \\
\bottomrule
\end{tabular}
\caption{Incremental ablation of selector training and bounded reflection on MMLongBench-Doc.
Each row adds the indicated component to the preceding configuration within the same routing block.
Selector F1 and average selected refer to pages in the page block and regions in the region block.}
\label{tab:selector-training-reflection}
\end{table}

\paragraph{Evidence Retrieval Evaluation}

To assess the page selector independently of downstream answering, we conduct a dedicated page-retrieval evaluation.
This protocol enables direct comparison with the page-selection components of prior methods and characterizes the quality of the evidence pool passed to later stages.
Such evaluation is critical because all subsequent region-selection experiments in HierDoc operate exclusively on the selected pages; page-level evidence omitted at this stage cannot be recovered downstream.
Table~\ref{tab:page-retrieval} reports page-level evidence precision, recall, F1, and the average number of selected pages.

The Doc-$V^*$ action rows describe the pages returned by two different tools rather than a single final page set, so they serve as reported reference points rather than strictly matched one-shot baselines.
Relative to the strongest reported Page-F1 reference on each benchmark, the HierDoc page policy improves Page F1 by 38.63\%, 56.76\%, and 12.89\% on MMLongBench-Doc, LongDocURL, and SlideVQA, respectively.
On MMLongBench-Doc, it additionally reduces the average number of selected pages by 60.71\% compared with Doc-$V^*$ with ColQwen.
Taken together, these gains demonstrate that the HierDoc page selector achieves excellent retrieval quality while constructing compact evidence sets, providing a reliable foundation for subsequent region selection and grounded answering.

\subsection{Ablation Studies}

\paragraph{Effects of Selector Training and Reflection}
Table~\ref{tab:selector-training-reflection} isolates the effects of policy training and bounded reflection at the two routing stages on MMLongBench-Doc.
The two incremental blocks make the contribution of each added component explicit: the page-policy ablation disables region routing, whereas the region-stage ablation evaluates region selection on top of the selected-page evidence.

For page selection, GRPO improves Page F1 by 18.32\% and downstream QA accuracy/F1 by 5.26\%/4.40\% relative to the untrained backbone.
Bounded reflection provides further relative gains of 1.62\% in Page F1 and 1.60\%/1.11\% in QA accuracy/F1, while reducing the average number of selected pages by 15.38\%.
These results indicate that policy training supplies the main localization improvement, whereas reflection removes redundant evidence without sacrificing downstream quality.

At the region stage, GRPO improves QA accuracy/F1 by 1.54\%/2.40\% over the untrained region policy even though Region F1 remains unchanged.
Adding bounded reflection then raises Region F1 by 26.87\%, improves QA accuracy/F1 by another 1.71\%/1.92\%, and reduces the average number of selected regions by 21.62\%.
Relative to page-only answering, the complete region stage improves QA accuracy/F1 by 5.51\%/4.82\%, demonstrating that trained region routing and reflection provide complementary gains beyond page selection.

\begin{figure*}[t]
\centering
\includegraphics[width=\textwidth]{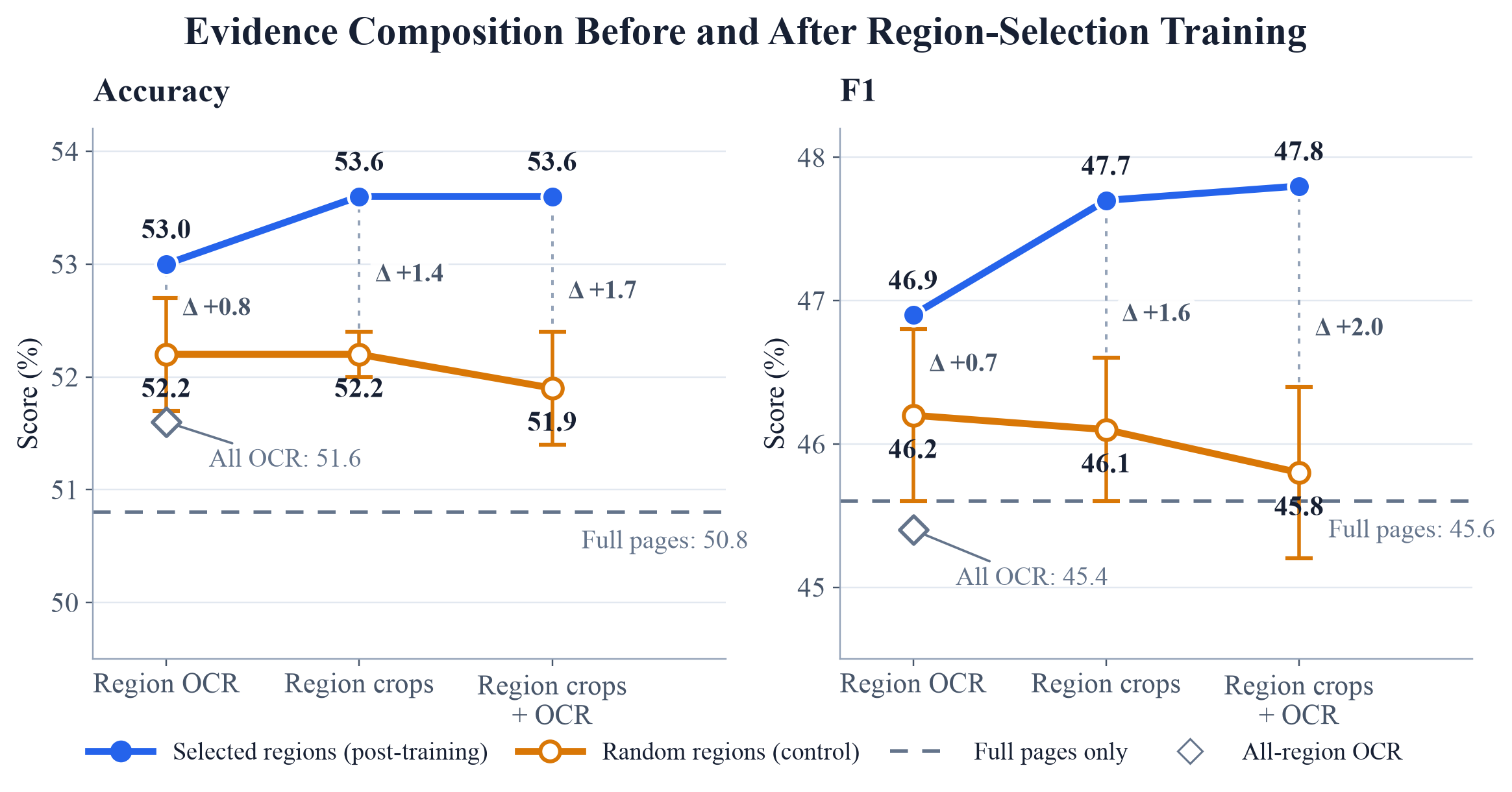}
\caption{Evidence-composition ablation for the final answer model. Random-region controls report the mean and sample standard deviation over four seeds (0--3), while all other points are single runs. The dashed line marks full-page-only performance, diamonds mark the all-region OCR baseline, and $\Delta$ denotes the absolute gain of trained region selection over the matched random-region control.}
\label{fig:evidence-composition}
\end{figure*}

\paragraph{Effects of Evidence Composition}
We further ablate the visual and textual evidence passed to the answer model while holding the selected pages and regions fixed.
Figure~\ref{fig:evidence-composition} compares full-page context, region crops, and OCR at different levels of selection.


Relative to full selected pages alone, appending OCR from every parsed region improves accuracy by 1.57\% but reduces F1 by 0.44\%, indicating that unfiltered text can introduce distracting context.
Across four random seeds, the matched random-region controls provide modest gains over page-only evidence: random crops improve accuracy/F1 by 2.85\%/1.17\%, random OCR by 2.79\%/1.28\%, and their combination by only 2.26\%/0.35\%.
Restricting OCR to selected regions instead yields relative gains of 4.33\% in accuracy and 2.85\% in F1.
Selected region crops provide larger gains of 5.51\%/4.61\%, and combining them with selected OCR produces the strongest configuration, improving accuracy/F1 by 5.51\%/4.82\% over page-only evidence.
More importantly, learned selection consistently outperforms the four-seed mean of its matched random controls: selected OCR improves accuracy/F1 by 1.50\%/1.55\%, selected crops by 2.59\%/3.39\%, and selected crops with OCR by 3.18\%/4.46\%.
In contrast, removing the full pages and retaining only selected crops and OCR reduces accuracy/F1 by 5.51\%/5.92\%.
Together, these results show that the gains arise primarily from question-conditioned region routing rather than merely adding more inputs; global page context and selected local evidence remain complementary, while targeted OCR provides a smaller additional refinement.

\section{Conclusion}

We introduced HierDoc, a hierarchical evidence-routing framework that formulates page and semantic-region acquisition as successive, answer-agnostic structured-set decisions.
HierDoc independently optimizes its page and region policies with stage-specific GRPO rewards, uses parser-native semantic regions as a discrete action space, and combines selected full pages, region crops, and textual metadata for grounded answering.
Across five multi-page and long-document VQA benchmarks, HierDoc achieves the best reported performance among open-weight methods on four benchmarks and competitive performance on the remaining one, with consistent improvements across different answer backbones.
The retrieval evaluation and controlled ablations further show that policy training substantially improves evidence localization, bounded reflection produces more compact evidence sets, and selected regional evidence complements the global context retained by full pages.
These findings demonstrate the effectiveness of organizing coarse-to-fine document understanding as an explicit and independently analyzable page-to-region evidence-routing process.

Several limitations remain.
First, the hierarchical design introduces irreversible error propagation: because the region policy operates only on the selected pages, evidence omitted during page routing cannot be recovered downstream.
Second, region routing depends on MinerU for candidate regions and textual metadata, so layout-parsing errors, OCR noise, and heuristic box-to-region alignment can constrain the available action space.
Third, the routing policies are optimized independently with set-level objectives rather than end-to-end answer feedback; consequently, improvements in evidence-selection metrics may not always translate proportionally into answer quality.
Future work could incorporate uncertainty-aware recovery across routing stages, jointly optimize evidence acquisition with grounded answering, and evaluate the resulting accuracy--efficiency trade-offs across broader document domains and languages.

\bibliography{references}

\begin{thebibliography}{37}
\providecommand{\natexlab}[1]{#1}

\bibitem[{Bai et~al.(2025{\natexlab{a}})Bai, Cai, Chen, Chen, Chen, Cheng,
  Deng, Ding, Gao, Ge, Ge, Guo, Huang, Huang, Huang, Hui, Jiang, Li, Li, Li,
  Li, Lin, Lin, Liu, Liu, Liu, Liu, Liu, Liu, Lu, Luo, Lv, Men, Meng, Ren, Ren,
  Song, Sun, Tang, Tu, Wan, Wang, Wang, Wang, Wang, Xie, Xu, Xu, Xu, Yang,
  Yang, Yang, Yang, Yu, Zhang, Zhang, Zhang, Zheng, Zhong, Zhou, Zhou, Zhou,
  Zhu, and Zhu}]{bai2025qwen3vl}
Bai, S.; Cai, Y.; Chen, R.; Chen, K.; Chen, X.; Cheng, Z.; Deng, L.; Ding, W.;
  Gao, C.; Ge, C.; Ge, W.; Guo, Z.; Huang, Q.; Huang, J.; Huang, F.; Hui, B.;
  Jiang, S.; Li, Z.; Li, M.; Li, M.; Li, K.; Lin, Z.; Lin, J.; Liu, X.; Liu,
  J.; Liu, C.; Liu, Y.; Liu, D.; Liu, S.; Lu, D.; Luo, R.; Lv, C.; Men, R.;
  Meng, L.; Ren, X.; Ren, X.; Song, S.; Sun, Y.; Tang, J.; Tu, J.; Wan, J.;
  Wang, P.; Wang, P.; Wang, Q.; Wang, Y.; Xie, T.; Xu, Y.; Xu, H.; Xu, J.;
  Yang, Z.; Yang, M.; Yang, J.; Yang, A.; Yu, B.; Zhang, F.; Zhang, H.; Zhang,
  X.; Zheng, B.; Zhong, H.; Zhou, J.; Zhou, F.; Zhou, J.; Zhu, Y.; and Zhu, K.
  2025{\natexlab{a}}.
\newblock {Qwen3-VL} Technical Report.
\newblock arXiv:2511.21631.

\bibitem[{Bai et~al.(2025{\natexlab{b}})Bai, Chen, Liu, Wang, Ge, Song, Dang,
  Wang, Wang, Tang, Zhong, Zhu, Yang, Li, Wan, Wang, Ding, Fu, Xu, Ye, Zhang,
  Xie, Cheng, Zhang, Yang, Xu, and Lin}]{bai2025qwen25vl}
Bai, S.; Chen, K.; Liu, X.; Wang, J.; Ge, W.; Song, S.; Dang, K.; Wang, P.;
  Wang, S.; Tang, J.; Zhong, H.; Zhu, Y.; Yang, M.; Li, Z.; Wan, J.; Wang, P.;
  Ding, W.; Fu, Z.; Xu, Y.; Ye, J.; Zhang, X.; Xie, T.; Cheng, Z.; Zhang, H.;
  Yang, Z.; Xu, H.; and Lin, J. 2025{\natexlab{b}}.
\newblock {Qwen2.5-VL} Technical Report.
\newblock arXiv:2502.13923.

\bibitem[{Chen et~al.(2025)Chen, Zhang, Zhou, Yu, Dernoncourt, Gu, Rossi, Chen,
  and Sun}]{chen2025svrag}
Chen, J.; Zhang, R.; Zhou, Y.; Yu, T.; Dernoncourt, F.; Gu, J.; Rossi, R.~A.;
  Chen, C.; and Sun, T. 2025.
\newblock {SV-RAG}: {LoRA}-Contextualizing Adaptation of {MLLMs} for Long
  Document Understanding.
\newblock In \emph{The Thirteenth International Conference on Learning
  Representations}.

\bibitem[{Cho et~al.(2024)Cho, Mahata, Irsoy, He, and Bansal}]{cho2024m3docrag}
Cho, J.; Mahata, D.; Irsoy, O.; He, Y.; and Bansal, M. 2024.
\newblock {M3DocRAG}: Multi-modal Retrieval is What You Need for Multi-page
  Multi-document Understanding.
\newblock arXiv:2411.04952.

\bibitem[{Deng et~al.(2025)Deng, Yuan, Bu, Wang, Li, Xu, Li, Gao, Song, Zheng,
  and Liu}]{deng-etal-2025-longdocurl}
Deng, C.; Yuan, J.; Bu, P.; Wang, P.; Li, Z.-Z.; Xu, J.; Li, X.-H.; Gao, Y.;
  Song, J.; Zheng, B.; and Liu, C.-L. 2025.
\newblock {LongDocURL}: A Comprehensive Multimodal Long Document Benchmark
  Integrating Understanding, Reasoning, and Locating.
\newblock In \emph{Proceedings of the 63rd Annual Meeting of the Association
  for Computational Linguistics (Volume 1: Long Papers)}, 1135--1159.

\bibitem[{Dong et~al.(2025)Dong, Chang, Goh Xin~Deik, Li, Tang, and
  Liu}]{dong-etal-2025-mmdocir}
Dong, K.; Chang, Y.; Goh Xin~Deik, D.; Li, D.; Tang, R.; and Liu, Y. 2025.
\newblock {MMDocIR}: Benchmarking Multimodal Retrieval for Long Documents.
\newblock In \emph{Proceedings of the 2025 Conference on Empirical Methods in
  Natural Language Processing}, 30971--31005.

\bibitem[{Faysse et~al.(2025)Faysse, Sibille, Wu, Omrani, Viaud, Hudelot, and
  Colombo}]{faysse2025colpali}
Faysse, M.; Sibille, H.; Wu, T.; Omrani, B.; Viaud, G.; Hudelot, C.; and
  Colombo, P. 2025.
\newblock {ColPali}: Efficient Document Retrieval with Vision Language Models.
\newblock In \emph{The Thirteenth International Conference on Learning
  Representations}.

\bibitem[{{GLM-V Team}(2025)}]{glm-v2025glm45v}
{GLM-V Team}. 2025.
\newblock {GLM-4.5V and GLM-4.1V-Thinking}: Towards Versatile Multimodal
  Reasoning with Scalable Reinforcement Learning.
\newblock arXiv:2507.01006.

\bibitem[{Han et~al.(2025)Han, Xia, Zhang, Sun, Li, Zhu, and
  Yao}]{han2025mdocagent}
Han, S.; Xia, P.; Zhang, R.; Sun, T.; Li, Y.; Zhu, H.; and Yao, H. 2025.
\newblock {MDocAgent}: A Multi-Modal Multi-Agent Framework for Document
  Understanding.
\newblock arXiv:2503.13964.

\bibitem[{Hu et~al.(2025)Hu, Xu, Zhang, Ye, Yan, Zhang, Jin, Huang, and
  Zhou}]{hu-etal-2025-mplug}
Hu, A.; Xu, H.; Zhang, L.; Ye, J.; Yan, M.; Zhang, J.; Jin, Q.; Huang, F.; and
  Zhou, J. 2025.
\newblock {mPLUG-DocOwl2}: High-resolution Compressing for {OCR}-free
  Multi-page Document Understanding.
\newblock In \emph{Proceedings of the 63rd Annual Meeting of the Association
  for Computational Linguistics (Volume 1: Long Papers)}, 5817--5834.

\bibitem[{Hui, Lu, and Zhang(2024)}]{hui2024uda}
Hui, Y.; Lu, Y.; and Zhang, H. 2024.
\newblock {UDA}: A Benchmark Suite for Retrieval Augmented Generation in
  Real-World Document Analysis.
\newblock In \emph{Advances in Neural Information Processing Systems},
  volume~37, 67200--67217.

\bibitem[{Li et~al.(2026)Li, Lu, Liu, Sun, Liu, and Xie}]{li2026regionrag}
Li, Y.; Lu, Z.; Liu, Z.; Sun, Y.; Liu, C.; and Xie, H. 2026.
\newblock {RegionRAG}: Region-level Retrieval-Augmented Generation for Visual
  Document Understanding.
\newblock \emph{Proceedings of the AAAI Conference on Artificial Intelligence},
  40(8): 6662--6670.

\bibitem[{Loison et~al.(2026)Loison, Mac{\'e}, Edy, Xing, Balough, Moreira,
  Liu, Faysse, Hudelot, and Viaud}]{loison2026vidore}
Loison, A.; Mac{\'e}, Q.; Edy, A.; Xing, V.; Balough, T.; Moreira, G.; Liu, B.;
  Faysse, M.; Hudelot, C.; and Viaud, G. 2026.
\newblock {ViDoRe V3}: A Comprehensive Evaluation of Retrieval Augmented
  Generation in Complex Real-World Scenarios.
\newblock arXiv:2601.08620.

\bibitem[{Ma et~al.(2024)Ma, Zang, Chen, Chen, Jiao, Li, Lu, Liu, Ma, Dong,
  Zhang, Pan, Jiang, Wang, Cao, and Sun}]{ma2024mmlongbenchdoc}
Ma, Y.; Zang, Y.; Chen, L.; Chen, M.; Jiao, Y.; Li, X.; Lu, X.; Liu, Z.; Ma,
  Y.; Dong, X.; Zhang, P.; Pan, L.; Jiang, Y.-G.; Wang, J.; Cao, Y.; and Sun,
  A. 2024.
\newblock {MMLongBench-Doc}: Benchmarking Long-context Document Understanding
  with Visualizations.
\newblock In \emph{Advances in Neural Information Processing Systems},
  volume~37, 95963--96010.

\bibitem[{Masry et~al.(2022)Masry, Long, Tan, Joty, and
  Hoque}]{masry-etal-2022-chartqa}
Masry, A.; Long, D.~X.; Tan, J.~Q.; Joty, S.; and Hoque, E. 2022.
\newblock {ChartQA}: A Benchmark for Question Answering about Charts with
  Visual and Logical Reasoning.
\newblock In \emph{Findings of the Association for Computational Linguistics:
  ACL 2022}, 2263--2279.

\bibitem[{Mathew et~al.(2022)Mathew, Bagal, Tito, Karatzas, Valveny, and
  Jawahar}]{Mathew_2022_WACV}
Mathew, M.; Bagal, V.; Tito, R.; Karatzas, D.; Valveny, E.; and Jawahar, C.
  2022.
\newblock InfographicVQA.
\newblock In \emph{Proceedings of the IEEE/CVF Winter Conference on
  Applications of Computer Vision (WACV)}, 1697--1706.

\bibitem[{Mathew, Karatzas, and Jawahar(2021)}]{Mathew_2021_WACV}
Mathew, M.; Karatzas, D.; and Jawahar, C. 2021.
\newblock DocVQA: A Dataset for VQA on Document Images.
\newblock In \emph{Proceedings of the IEEE/CVF Winter Conference on
  Applications of Computer Vision (WACV)}, 2200--2209.

\bibitem[{Niu et~al.(2026)Niu, Liu, Gu, Wang, Ouyang, Zhao, Chu, He, Wu, Zhang,
  Jin, Liang, Zhang, Zhang, Qu, Ren, Sun, Tang, Niu, Zheng, Ma, Miao, Dong,
  Qian, Zhang, Wang, Chen, Zhao, Wei, Li, Wang, Xu, Cao, Chen, Wu, Gu, Lu, Lin,
  Shen, Zhou, Zhang, Zang, Dong, Wang, Zhang, Bai, Chu, Li, Wu, Wu, Li, Wang,
  Tu, Xu, Chen, Zhou, Lin, Zhang, and He}]{niu-etal-2026-mineru2}
Niu, J.; Liu, Z.; Gu, Z.; Wang, B.; Ouyang, L.; Zhao, Z.; Chu, T.; He, T.; Wu,
  F.; Zhang, Q.; Jin, Z.; Liang, G.; Zhang, R.; Zhang, W.; Qu, Y.; Ren, Z.;
  Sun, Y.; Tang, Z.; Niu, B.; Zheng, Y.; Ma, D.; Miao, Z.; Dong, H.; Qian, S.;
  Zhang, J.; Wang, F.; Chen, J.; Zhao, X.; Wei, L.; Li, W.; Wang, S.; Xu, R.;
  Cao, Y.; Chen, L.; Wu, Q.; Gu, H.; Lu, L.; Lin, D.; Shen, G.; Zhou, X.;
  Zhang, L.; Zang, Y.; Dong, X.; Wang, J.; Zhang, B.; Bai, L.; Chu, P.; Li, W.;
  Wu, J.; Wu, L.; Li, Z.; Wang, G.; Tu, Z.; Xu, C.; Chen, K.; Zhou, B.; Lin,
  D.; Zhang, W.; and He, C. 2026.
\newblock {MinerU2.5}: A Decoupled Vision-Language Model for Efficient
  High-Resolution Document Parsing.
\newblock In \emph{Proceedings of the 64th Annual Meeting of the Association
  for Computational Linguistics (Volume 6: Industry Track)}, 13--42.

\bibitem[{Shao et~al.(2024)Shao, Wang, Zhu, Xu, Song, Bi, Zhang, Zhang, Li, Wu,
  and Guo}]{shao2024deepseekmath}
Shao, Z.; Wang, P.; Zhu, Q.; Xu, R.; Song, J.; Bi, X.; Zhang, H.; Zhang, M.;
  Li, Y.~K.; Wu, Y.; and Guo, D. 2024.
\newblock {DeepSeekMath}: Pushing the Limits of Mathematical Reasoning in Open
  Language Models.
\newblock arXiv:2402.03300.

\bibitem[{Tanaka et~al.(2023)Tanaka, Nishida, Nishida, Hasegawa, Saito, and
  Saito}]{tanaka2023slidevqa}
Tanaka, R.; Nishida, K.; Nishida, K.; Hasegawa, T.; Saito, I.; and Saito, K.
  2023.
\newblock {SlideVQA}: A Dataset for Document Visual Question Answering on
  Multiple Images.
\newblock \emph{Proceedings of the AAAI Conference on Artificial Intelligence},
  37(11): 13636--13645.

\bibitem[{Tito, Karatzas, and Valveny(2023)}]{tito2023hierarchical}
Tito, R.; Karatzas, D.; and Valveny, E. 2023.
\newblock Hierarchical Multimodal Transformers for Multi-Page DocVQA.
\newblock \emph{Pattern Recognition}, 144: 109834.

\bibitem[{Van~Landeghem et~al.(2023)Van~Landeghem, Tito, Borchmann, Pietruszka,
  Joziak, Powalski, Jurkiewicz, Coustaty, Anckaert, Valveny, Blaschko, Moens,
  and Stanislawek}]{Van_Landeghem_2023_ICCV}
Van~Landeghem, J.; Tito, R.; Borchmann, {\L}.; Pietruszka, M.; Joziak, P.;
  Powalski, R.; Jurkiewicz, D.; Coustaty, M.; Anckaert, B.; Valveny, E.;
  Blaschko, M.; Moens, S.; and Stanislawek, T. 2023.
\newblock Document Understanding Dataset and Evaluation (DUDE).
\newblock In \emph{Proceedings of the IEEE/CVF International Conference on
  Computer Vision (ICCV)}, 19528--19540.

\bibitem[{Wang et~al.(2025{\natexlab{a}})Wang, Ding, Chen, Wu, Wang, Xie, and
  Zhao}]{wang-etal-2025-vidorag}
Wang, Q.; Ding, R.; Chen, Z.; Wu, W.; Wang, S.; Xie, P.; and Zhao, F.
  2025{\natexlab{a}}.
\newblock {ViDoRAG}: Visual Document Retrieval-Augmented Generation via Dynamic
  Iterative Reasoning Agents.
\newblock In \emph{Proceedings of the 2025 Conference on Empirical Methods in
  Natural Language Processing}, 9113--9134.

\bibitem[{Wang et~al.(2025{\natexlab{b}})Wang, Ding, Zeng, Chen, Chen, Wang,
  Xie, Huang, and Zhao}]{wang2025vragrl}
Wang, Q.; Ding, R.; Zeng, Y.; Chen, Z.; Chen, L.; Wang, S.; Xie, P.; Huang, F.;
  and Zhao, F. 2025{\natexlab{b}}.
\newblock {VRAG-RL}: Empower Vision-Perception-Based {RAG} for Visually Rich
  Information Understanding via Iterative Reasoning with Reinforcement
  Learning.
\newblock arXiv:2505.22019.

\bibitem[{Wang et~al.(2025{\natexlab{c}})Wang, Gao, Gu, Pu, Cui, Wei, Liu,
  Jing, Ye, Shao, Wang, Chen, Zhang, Yang, Wang, Wei, Yin, Li, Cui, Chen, Ding,
  Tian, Wu, Xie, Li, Yang, Duan, Wang, Hou, Hao, Zhang, Li, Zhao, Duan, Deng,
  Fu, He, Wang, He, Shi, He, Xiong, Lv, Wu, Shao, Zhang, Deng, Qi, Ge, Guo,
  Zhang, Zhang, Cao, Lin, Tang, Gao, Huang, Gu, Lyu, Tang, Wang, Lv, Ouyang,
  Wang, Dou, Zhu, Lu, Lin, Dai, Su, Zhou, Chen, Qiao, Wang, and
  Luo}]{wang2025internvl35}
Wang, W.; Gao, Z.; Gu, L.; Pu, H.; Cui, L.; Wei, X.; Liu, Z.; Jing, L.; Ye, S.;
  Shao, J.; Wang, Z.; Chen, Z.; Zhang, H.; Yang, G.; Wang, H.; Wei, Q.; Yin,
  J.; Li, W.; Cui, E.; Chen, G.; Ding, Z.; Tian, C.; Wu, Z.; Xie, J.; Li, Z.;
  Yang, B.; Duan, Y.; Wang, X.; Hou, Z.; Hao, H.; Zhang, T.; Li, S.; Zhao, X.;
  Duan, H.; Deng, N.; Fu, B.; He, Y.; Wang, Y.; He, C.; Shi, B.; He, J.; Xiong,
  Y.; Lv, H.; Wu, L.; Shao, W.; Zhang, K.; Deng, H.; Qi, B.; Ge, J.; Guo, Q.;
  Zhang, W.; Zhang, S.; Cao, M.; Lin, J.; Tang, K.; Gao, J.; Huang, H.; Gu, Y.;
  Lyu, C.; Tang, H.; Wang, R.; Lv, H.; Ouyang, W.; Wang, L.; Dou, M.; Zhu, X.;
  Lu, T.; Lin, D.; Dai, J.; Su, W.; Zhou, B.; Chen, K.; Qiao, Y.; Wang, W.; and
  Luo, G. 2025{\natexlab{c}}.
\newblock {InternVL3.5}: Advancing Open-Source Multimodal Models in
  Versatility, Reasoning, and Efficiency.
\newblock arXiv:2508.18265.

\bibitem[{Wang et~al.(2026)Wang, Ma, Zhong, Li, Zhang, Wang, and
  He}]{wang2026agenticocr}
Wang, Z.; Ma, D.; Zhong, H.; Li, J.; Zhang, W.; Wang, B.; and He, C. 2026.
\newblock {AgenticOCR}: Parsing Only What You Need for Efficient
  Retrieval-Augmented Generation.
\newblock arXiv:2602.24134.

\bibitem[{Wei, Sun, and Li(2025)}]{wei2025deepseekocr}
Wei, H.; Sun, Y.; and Li, Y. 2025.
\newblock {DeepSeek-OCR}: Contexts Optical Compression.
\newblock arXiv:2510.18234.

\bibitem[{Wu et~al.(2025)Wu, Tan, Hou, Zhang, and Cheng}]{wu-etal-2025-molorag}
Wu, X.; Tan, Y.; Hou, N.; Zhang, R.; and Cheng, H. 2025.
\newblock {MoLoRAG}: Bootstrapping Document Understanding via Multi-modal
  Logic-aware Retrieval.
\newblock In \emph{Proceedings of the 2025 Conference on Empirical Methods in
  Natural Language Processing}, 14024--14045.

\bibitem[{Xiong et~al.(2026{\natexlab{a}})Xiong, Wang, Gu, Liu, Yin, Zhou, and
  Li}]{xiong2026hierarchical}
Xiong, J.; Wang, Y.; Gu, R.; Liu, C.; Yin, B.; Zhou, W.; and Li, H.
  2026{\natexlab{a}}.
\newblock Hierarchical Evidence-Driven Reasoning for Long Document
  Understanding.
\newblock arXiv:2607.04625.

\bibitem[{Xiong et~al.(2026{\natexlab{b}})Xiong, Wang, Zhao, Liu, Yin, Zhou,
  and Li}]{xiong2026docr1}
Xiong, J.; Wang, Y.; Zhao, W.; Liu, C.; Yin, B.; Zhou, W.; and Li, H.
  2026{\natexlab{b}}.
\newblock {DocR1}: Evidence Page-Guided {GRPO} for Multi-Page Document
  Understanding.
\newblock \emph{Proceedings of the AAAI Conference on Artificial Intelligence},
  40(13): 11178--11186.

\bibitem[{Xu et~al.(2025)Xu, Wang, Liu, Lin, and Chen}]{xu2025cogdoc}
Xu, Q.; Wang, H.; Liu, C.; Lin, F.; and Chen, W. 2025.
\newblock {CogDoc}: Towards Unified thinking in Documents.
\newblock arXiv:2512.12658.

\bibitem[{Yu et~al.(2025)Yu, Tang, Xu, Cui, Ran, Yan, Liu, Wang, Han, Liu, and
  Sun}]{yu2025visrag}
Yu, S.; Tang, C.; Xu, B.; Cui, J.; Ran, J.; Yan, Y.; Liu, Z.; Wang, S.; Han,
  X.; Liu, Z.; and Sun, M. 2025.
\newblock {VisRAG}: Vision-based Retrieval-augmented Generation on
  Multi-modality Documents.
\newblock In \emph{The Thirteenth International Conference on Learning
  Representations}.

\bibitem[{Zheng et~al.(2023)Zheng, Chiang, Sheng, Zhuang, Wu, Zhuang, Lin, Li,
  Li, Xing, Zhang, Gonzalez, and Stoica}]{zheng2023judging}
Zheng, L.; Chiang, W.-L.; Sheng, Y.; Zhuang, S.; Wu, Z.; Zhuang, Y.; Lin, Z.;
  Li, Z.; Li, D.; Xing, E.~P.; Zhang, H.; Gonzalez, J.~E.; and Stoica, I. 2023.
\newblock Judging {LLM}-as-a-Judge with {MT-Bench} and {Chatbot Arena}.
\newblock In \emph{Advances in Neural Information Processing Systems},
  volume~36, 46595--46623.

\bibitem[{Zheng et~al.(2026)Zheng, Fu, Li, Wang, Zhang, Ruan, Zhang, Wei, Luo,
  Luan, Chen, and Bai}]{zheng-etal-2026-doc}
Zheng, Y.; Fu, P.; Li, H.; Wang, Z.; Zhang, Y.; Ruan, W.; Zhang, X.; Wei, Z.;
  Luo, Z.; Luan, J.; Chen, W.; and Bai, X. 2026.
\newblock Doc-$V^*$: Coarse-to-Fine Interactive Visual Reasoning for Multi-Page
  Document VQA.
\newblock In \emph{Proceedings of the 64th Annual Meeting of the Association
  for Computational Linguistics (Volume 1: Long Papers)}, 45901--45923.

\bibitem[{Zhu et~al.(2026)Zhu, Meng, Chen, Li, Pfister, and
  Yoon}]{zhu-etal-2026-doclens}
Zhu, D.; Meng, R.; Chen, J.; Li, S.; Pfister, T.; and Yoon, J. 2026.
\newblock {DocLens}: A Tool-Augmented Multi-Agent Framework for Long Visual
  Document Understanding.
\newblock In \emph{Proceedings of the 64th Annual Meeting of the Association
  for Computational Linguistics (Volume 1: Long Papers)}, 26804--26829.

\bibitem[{Zhu et~al.(2022)Zhu, Lei, Feng, Wang, Zhang, and
  Chua}]{zhu2022tatdqa}
Zhu, F.; Lei, W.; Feng, F.; Wang, C.; Zhang, H.; and Chua, T.-S. 2022.
\newblock Towards Complex Document Understanding By Discrete Reasoning.
\newblock In \emph{Proceedings of the 30th ACM International Conference on
  Multimedia}, 4857--4866.

\bibitem[{Zhu et~al.(2025)Zhu, Wang, Chen, Liu, Ye, Gu, Tian, Duan, Su, Shao,
  Gao, Cui, Wang, Cao, Liu, Wei, Zhang, Wang, Xu, Li, Wang, Deng, Li, He,
  Jiang, Luo, Wang, He, Shi, Zhang, Shao, He, Xiong, Qu, Sun, Jiao, Lv, Wu,
  Zhang, Deng, Ge, Chen, Wang, Dou, Lu, Zhu, Lu, Lin, Qiao, Dai, and
  Wang}]{zhu2025internvl3}
Zhu, J.; Wang, W.; Chen, Z.; Liu, Z.; Ye, S.; Gu, L.; Tian, H.; Duan, Y.; Su,
  W.; Shao, J.; Gao, Z.; Cui, E.; Wang, X.; Cao, Y.; Liu, Y.; Wei, X.; Zhang,
  H.; Wang, H.; Xu, W.; Li, H.; Wang, J.; Deng, N.; Li, S.; He, Y.; Jiang, T.;
  Luo, J.; Wang, Y.; He, C.; Shi, B.; Zhang, X.; Shao, W.; He, J.; Xiong, Y.;
  Qu, W.; Sun, P.; Jiao, P.; Lv, H.; Wu, L.; Zhang, K.; Deng, H.; Ge, J.; Chen,
  K.; Wang, L.; Dou, M.; Lu, L.; Zhu, X.; Lu, T.; Lin, D.; Qiao, Y.; Dai, J.;
  and Wang, W. 2025.
\newblock {InternVL3}: Exploring Advanced Training and Test-Time Recipes for
  Open-Source Multimodal Models.
\newblock arXiv:2504.10479.

\end{thebibliography}

\clearpage
\twocolumn[
\begin{center}
{\LARGE\bfseries Supplementary Material}
\end{center}
\vspace{1em}
]

This supplement reports dataset accounting, benchmark-specific evaluation choices, the GRPO objectives used for the two routing policies, complete inference prompts, additional ablations, and space for extended qualitative examples.
Unless stated otherwise, ``page'' denotes a 1-based physical rendered page rather than a printed PDF page label.

\appendix

\section{Datasets and Evaluation Protocols}
\label{supp:evaluation}

\subsection{Dataset Scale}

Table~\ref{tab:supp-dataset-scale} reports both the number of unique rendered page images and the number of page-image inputs accumulated over questions.
The latter repeats a document's pages whenever multiple questions refer to the same document and therefore reflects the uncompressed evaluation workload.
All counts are computed before page routing, window padding, or evidence selection.

\begin{table*}[t]
\centering%
\begin{tabular}{lrrrrr}
\toprule
Dataset & Samples & Documents / decks & Unique images & Accumulated image inputs \\
\midrule
MMLongBench-Doc~\citep{ma2024mmlongbenchdoc} & 1,082 & 135 & 6,529 & 51,699 \\
LongDocURL~\citep{deng-etal-2025-longdocurl} & 2,325 & 396 & 16,230 & 69,750 \\
SlideVQA~\citep{tanaka2023slidevqa} & 2,215 & 400 & 8,000 & 44,300 \\
PaperTab~\citep{hui2024uda} & 393 & 307 & 3,375 & 4,211 \\
FetaTab~\citep{hui2024uda} & 1,023 & 878 & 13,822 & 16,708 \\
\bottomrule
\end{tabular}%
\caption{Dataset image and sample counts used in our evaluation.
``Unique page images'' deduplicates physical pages within each evaluated split, whereas ``accumulated'' counts all pages supplied for every question before routing.}
\label{tab:supp-dataset-scale}
\end{table*}

\subsection{MMLongBench-Doc Dataset Size}
\label{supp:mmlong-cleanup}

The initial MMLongBench-Doc release contained 1,091 records, whereas the current released test file contains 1,082.
We use the current 1,082-record release.
The nine-record net reduction is an upstream data-cleaning change, not model-dependent filtering by our method.
Some affected entries are declarative statements rather than questions (e.g., ``there are four different rungs for the ladder of causation'').
They are marked \texttt{Not answerable} and have no annotated evidence page, so they do not define a grounded page-selection or VQA target.

\subsection{Semantic Evaluation for PaperTab and FetaTab}
\label{supp:uda-judge}

PaperTab and FetaTab use free-form, sentence-level answers rather than short extractive spans.
Token-level F1 is difficult to interpret in this setting because a semantically correct response may use different wording or a different level of detail from the reference.
We therefore use LLM-as-a-judge, a common scalable protocol for open-ended generation evaluation, with an explicit semantic-correctness rubric~\citep{zheng2023judging}.

This choice does not introduce an evaluation mismatch between HierDoc and the comparison systems. 
The other methods also report LLM-as-a-judge results on PaperTab and FetaTab for the same reason, and we use their reported scores directly.
Thus, the compared results consistently use LLM-based semantic evaluation rather than mixing token-F1 scores with judge scores.
Table~\ref{tab:supp-judge-examples} gives two representative cases from our outputs.

\begin{table*}[t]
\centering
\small
\begin{tabular}{p{0.10\textwidth}p{0.34\textwidth}p{0.26\textwidth}cc}
\toprule
Dataset & Reference & Prediction & Token F1 & Judge score \\
\midrule
FetaTab & In 2016, Platt played the role of Evan Hansen at Second Stage Theatre. & Evan Hansen & 0.286 & 1.0 \\
PaperTab & SimpleQuestions, WebQSP & SimpleQuestions and WebQSP & 0.800 & 1.0 \\
\bottomrule
\end{tabular}
\caption{Why lexical overlap alone can understate semantic correctness. The
first prediction supplies exactly the requested role; the second returns the
same two benchmarks with a natural-language list separator.}
\label{tab:supp-judge-examples}
\end{table*}

\section{Stage-Wise GRPO Training}
\label{supp:grpo}

\subsection{Algorithm Overview}

We independently optimize the page and region selectors with Group Relative Policy Optimization (GRPO)~\citep{shao2024deepseekmath}.
For each prompt $x$, the old policy samples a group of $G=4$ structured selections $\{o_i\}_{i=1}^{G}$.
Their scalar rewards are standardized within the group,

\begin{equation}
\widehat A_i =
\frac{R_i-\overline R}{\operatorname{Std}(R_1,\ldots,R_G)+\epsilon}.
\end{equation}

Writing
$\rho_{i,t}=\pi_\theta(o_{i,t}\mid x,o_{i,<t})/
\pi_{\mathrm{old}}(o_{i,t}\mid x,o_{i,<t})$ and
$\widetilde\rho_{i,t}=\operatorname{clip}
(\rho_{i,t},1-\varepsilon,1+\varepsilon)$, the update is

\begin{align}
\mathcal J_{\mathrm{GRPO}}(\theta)
={}&\frac{1}{G}\sum_{i=1}^{G}\frac{1}{|o_i|}
\sum_{t=1}^{|o_i|}\min\!\left(\rho_{i,t}\widehat A_i,
\widetilde\rho_{i,t}\widehat A_i\right) \nonumber\\
&-\beta D_{\mathrm{KL}}
(\pi_\theta\,\|\,\pi_{\mathrm{ref}}),
\end{align}

where the reference-policy coefficient is $\beta=0.01$.
Relative advantages remove the need for a learned value critic and compare alternative evidence sets for the same document-question context.
This is especially natural here: the action is a discrete set serialized in an XML tag, and its quality can be scored deterministically against annotated evidence.
\subsection{Stage-Specific Rewards}

Let $S$ and $G$ be the predicted and gold evidence sets. We compute set
recall, precision, and F1 after parsing the single required XML tag. The reward
weights are shown in Table~\ref{tab:supp-reward-weights}.

\begin{table}[t]
\centering
\begin{tabular}{lrrrr}
\toprule
Selector & Recall & F1 & Precision & Format \\
\midrule
Page & 0.40 & 0.45 & 0.00 & 0.15 \\
Region & 0.20 & 0.50 & 0.20 & 0.10 \\
\bottomrule
\end{tabular}
\caption{Coefficients of the selector rewards. Each row sums to one before
invalid-action penalties or caps.}
\label{tab:supp-reward-weights}
\end{table}

The page policy is intentionally recall-sensitive because a missed page is an irreversible error: the region selector cannot inspect a page it never receives.
Set F1 discourages indiscriminate page selection, while the format term teaches the exact \texttt{<evidence\_page>} protocol.
Invalid page IDs incur a penalty of $\min(0.20,0.05|U_p|)$, where $U_p$ is the set of IDs not present in the candidate window.

The region action space is denser, so its reward explicitly includes precision.
A missing region tag receives zero reward; a non-strict parse is capped at 0.25, and any invalid numeric alias caps the reward at 0.10.
All rewards are clipped to $[0,1]$.
This asymmetry encodes the cascade's error structure: page routing favors sufficient coverage, whereas region routing must retain that evidence while producing a compact crop set.

\FloatBarrier

\section{Additional Ablations}
\label{supp:ablations}

\subsection{Recall--F1 Reward Balance}

\begin{table*}[p]
\centering
\resizebox{\textwidth}{!}{%
\begin{tabular}{lccccccccc}
\toprule
Reward configuration & $\lambda_{\mathrm{F1}}^p$ & $\lambda_{\mathrm{rec}}^p$ & Macro P & Macro R & Macro F1 & Full coverage & Avg. pages & QA Acc. & QA F1 \\
\midrule
F1-based & 0.85 & 0.00 & 54.5 & 78.9 & 59.5 & 69.3 & 3.8 & 49.4 & 44.7 \\
\textbf{Balanced} & \textbf{0.45} & \textbf{0.40} & \textbf{66.3} & 77.4 & \textbf{67.8} & 67.8 & \textbf{2.6} & \textbf{50.0} & \textbf{45.1} \\
Recall-based & 0.10 & 0.75 & 63.4 & \textbf{79.0} & 65.8 & \textbf{69.9} & 3.1 & 49.6 & 44.6 \\
\bottomrule
\end{tabular}%
}
\caption{Ablation of the recall--F1 balance in the page-policy reward.}
\label{tab:reward-balance}
\end{table*}

We keep the page-format coefficient fixed at $\lambda_{\mathrm{fmt}}^p=0.15$ and vary the coverage--compactness balance.
The balanced reward gives the best downstream QA accuracy/F1 (49.97/45.09) while selecting only 2.645 pages on average.
The recall-heavy variant obtains the highest selector recall and full-evidence coverage (78.95 and 69.89), but expands the selected set to 3.123 pages and reduces downstream QA to 49.59/44.60.
Conversely, the F1-based setting selects 3.766 pages and is less precise.
Thus, maximizing one retrieval statistic in isolation is inferior to balancing evidence coverage with a compact input context.

\subsection{Same-Document Negative Pages}

\begin{table}[H]
\centering
\begin{tabular}{lcc}
\toprule
Region-policy training data & Accuracy & F1 \\
\midrule
Evidence pages only & 55.2 & 47.4 \\
\textbf{Evidence + negative pages} & \textbf{55.6} & \textbf{47.9} \\
\bottomrule
\end{tabular}
\caption{Effect of same-document negative pages in region-policy training.}
\label{tab:negative-pages}
\end{table}

With oracle evidence pages, identical parser candidates, training only on evidence pages yields 55.18 accuracy and 47.38 F1.
Adding non-evidence pages from the same document raises these scores to 55.62 and 47.89.
The gains of 0.44 accuracy and 0.51 F1 points show that document-specific distractors provide useful negative signal without changing the region policy's semantic action space.

\section{Additional Qualitative Examples}
\label{supp:qualitative}

\begin{figure*}[p]
\centering
\includegraphics[width=0.98\textwidth]{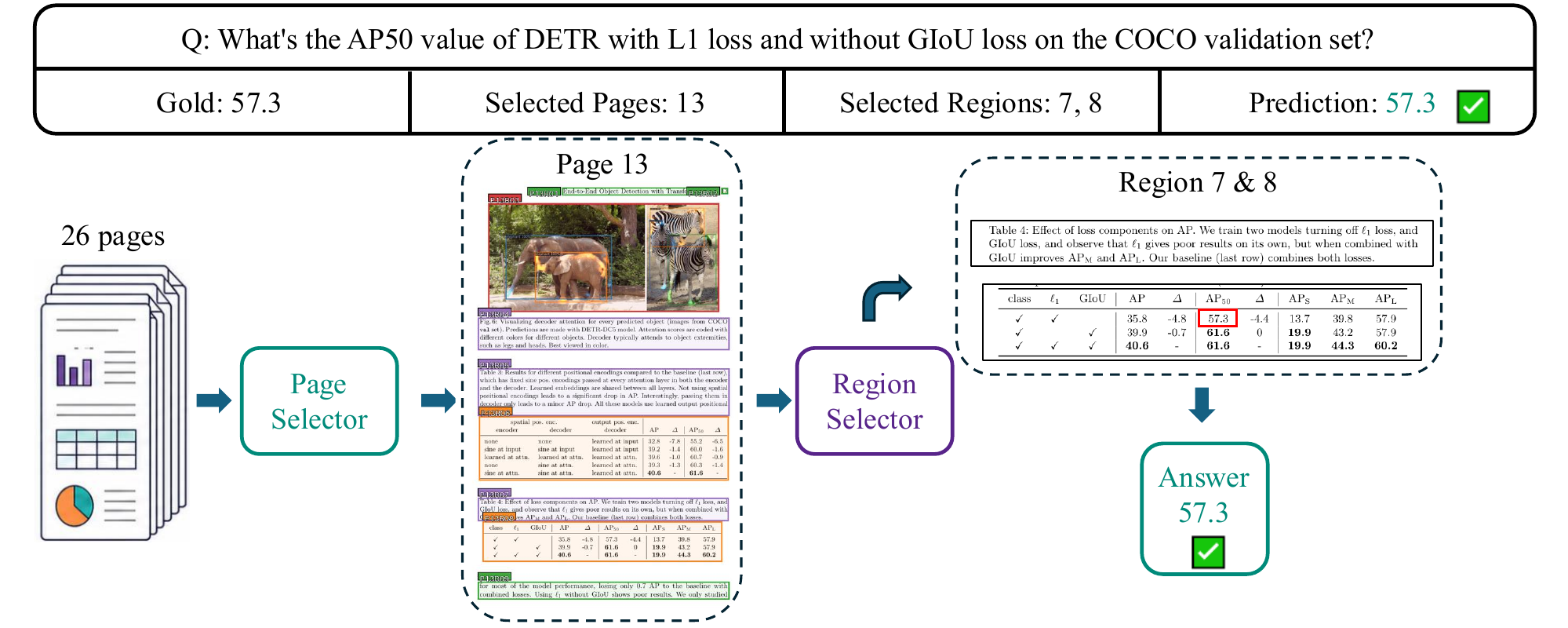}
\caption{Single-page evidence routing on MMLongBench-Doc. For an AP$_{50}$
lookup over a 26-page DETR paper, the page selector retains page~13. The
region selector then selects aliases 7 and 8, which provide the loss-component
caption and the corresponding ablation table. From this compact evidence, the
answer model returns 57.3, matching the reference answer.}
\label{fig:supp-single-page}

\vspace{0.5em}
\includegraphics[width=0.98\textwidth]{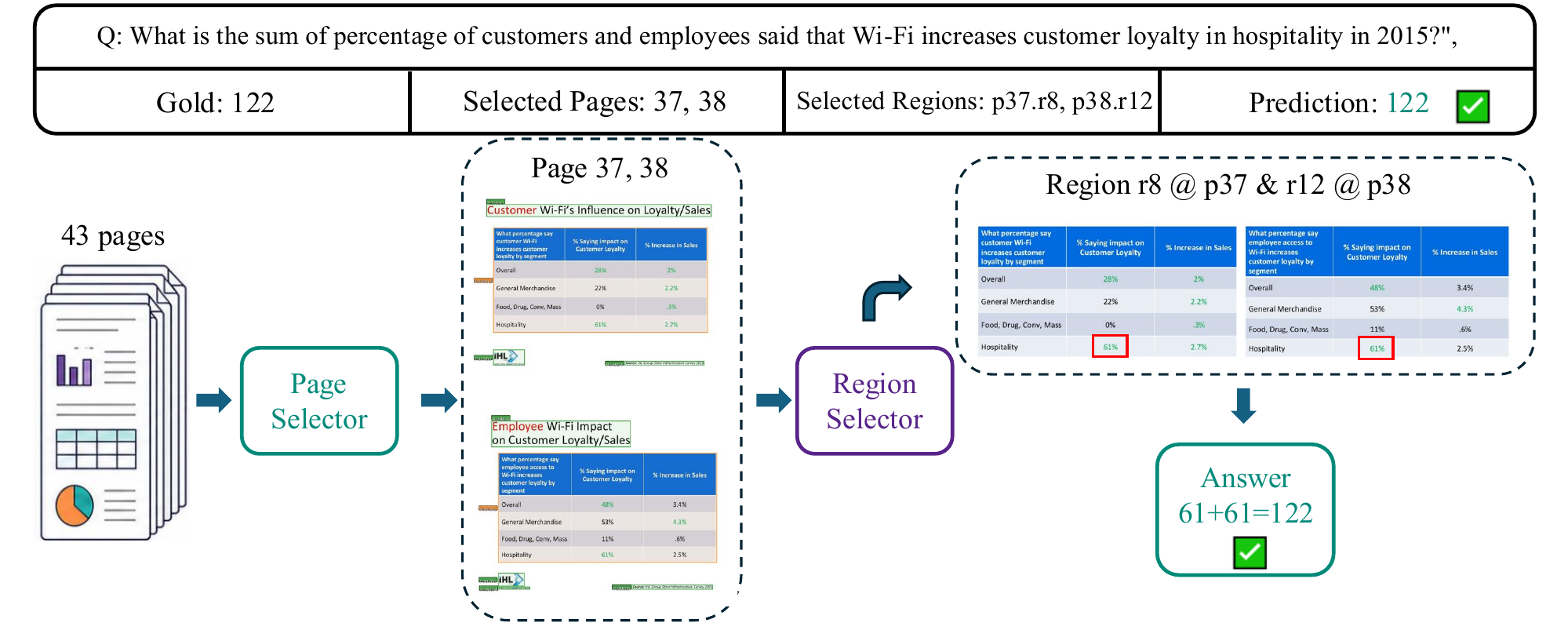}
\caption{Cross-page compositional routing on MMLongBench-Doc. From a 43-page
document, the page selector retains pages~37 and~38. The region selector
isolates the hospitality rows as p37.r8 and p38.r12, which respectively report
61\% for customer and employee Wi-Fi. The answer model aggregates the two
values as $61+61=122$, again matching the reference answer.}
\label{fig:supp-cross-page}
\end{figure*}

Figures~\ref{fig:supp-single-page} and~\ref{fig:supp-cross-page} illustrate
two successful routing patterns on MMLongBench-Doc. Both examples expose the
question, reference answer, selected page IDs, selected region aliases, and
final prediction, making the intermediate evidence decisions directly
auditable. The first example isolates a caption--table pair from a single
page, whereas the second combines values retrieved from two different pages.

\section{Complete Inference Prompts}
\label{supp:prompts}

Listings~\ref{lst:supp-page-prompt}--\ref{lst:supp-answer-prompt} show the three-stage prompts for page selection, region selection, and final answer fusion.
Braced fields and \texttt{<image>} denote values and multimodal image slots inserted at runtime.

\FloatBarrier

\begin{listing*}[t]
\begin{lstlisting}
You are an evidence-page selector for multi-page document VQA.

Task: given a question and the candidate document page images, select every candidate page that contains information needed to answer the question.

Rules:
- Use only the page numbers printed before the images.
- Do not answer the question.
- Do not output JSON or markdown.
- Return page numbers in ascending order, separated by commas.
- The final response must be exactly one tag like <evidence_page>2,3,4</evidence_page>.
- If no candidate page contains evidence, return <evidence_page></evidence_page>.

Format examples:
Candidate pages: 1,2,3,4
Question: Which pages show the revenue table and its footnote?
Output: <evidence_page>2,4</evidence_page>

Candidate pages: 5,8,9
Question: What is the invoice number?
Output: <evidence_page>5</evidence_page>

Now select the evidence pages for the real document.

Candidate pages: {candidate_page_numbers}

Candidate page {page_number}:
<image>
...

Question:
{question}

Output:
\end{lstlisting}
\caption{Stage 1: evidence-page selector prompt.}
\label{lst:supp-page-prompt}
\end{listing*}

\begin{listing*}[t]
\begin{lstlisting}
You are an evidence-region selector for document VQA.

Task: given a question, candidate page images, and candidate region IDs, select every candidate region needed to answer the question.

Rules:
- Region IDs are numeric aliases assigned only for this sample.
- Use only the listed numeric region IDs.
- Do not answer the question.
- Do not output JSON or markdown.
- Prioritize recall: include every candidate region that may be needed to answer the question.
- Missing required evidence is worse than including a few extra relevant or adjacent regions.
- Select a compact high-recall set; do not blindly select all candidates or clearly irrelevant regions.
- For tables, charts, figures, forms, captions, legends, axes, headers, footnotes, sources, comparisons, counts, aggregations, or cross-page evidence, include every supporting region needed to interpret the evidence.
- When uncertain between related neighboring regions, include the plausible evidence region.
- The candidate regions list gives each numeric ID, region type, and OCR hint.
- The page images are authoritative. OCR/table text is only a helpful hint.
- Return numeric region IDs in page/image order, separated by commas.
- The final response must be exactly one tag like <evidence_region>3,17</evidence_region>.
- If no candidate region contains evidence, return <evidence_region></evidence_region>.

Format examples:
Candidate region IDs: 1,2,3
Question: Which table shows total revenue?
Output: <evidence_region>2</evidence_region>

Candidate region IDs: 4,5,6,7
Question: Which chart and caption compare revenue and profit?
Output: <evidence_region>4,7</evidence_region>

Candidate region IDs: 8,9
Question: What is the invoice number?
Output: <evidence_region></evidence_region>

Now select the evidence regions for the real document.

Candidate pages and images:
{candidate_pages_and_images}

Candidate region IDs: {candidate_region_ids}

Candidate regions:
{numeric_alias} | type={region_type} | OCR/table text={ocr_hint}
...

Question:
{question}

Final reminder: represent the selected evidence regions only with one <evidence_region>...</evidence_region> tag. Do not output page IDs, JSON, markdown, or the answer.

Output:
\end{lstlisting}
\caption{Stage 2: evidence-region selector prompt.}
\label{lst:supp-region-prompt}
\end{listing*}

\begin{listing*}[t]
\begin{lstlisting}
[System]
You are a document visual question-answering model. You will receive selected full page images from one document plus cropped evidence regions and cleaned OCR/table text extracted from regions selected by an evidence-region selector.

Use only the provided page images, region crops, and selected OCR/table text to answer the question. The page and crop images are authoritative; the selected OCR/table text is a helpful hint about likely evidence. Combine information across pages and regions when needed and follow the requested answer format.

Return only the final answer. Do not include explanations, citations, page numbers, markdown, or extra text. If the expected answer is a list, return a valid JSON array. If the answer is not visible in the provided pages and selected OCR/table text, return exactly "Not answerable".

[User]
Selected page images:
Page {page_number}:
<image>
...

Selected evidence regions:
Region id={region_id}, page={page_number}, type={region_type}, bbox={x1,y1,x2,y2}
Crop image:
<image>
OCR/table text:
{cleaned_ocr_or_table_text}
...

Question:
{question}

Expected answer format: {answer_format}
Final answer:
\end{lstlisting}
\caption{Stage 3: final fusion-and-answer prompt. For PaperTab and FetaTab,
the system instruction additionally emphasizes academic-paper/Wikipedia
documents, exhaustive use of tables, captions, footnotes, formulas, and
cross-page context, and exact \texttt{Yes}/\texttt{No} output.}
\label{lst:supp-answer-prompt}
\end{listing*}

\end{document}